\documentclass[10pt, conference]{IEEEtran}
\IEEEoverridecommandlockouts
\usepackage{cite}
\usepackage{amsmath,amssymb,amsfonts}
\usepackage{algorithmic}
\usepackage{graphicx}
\usepackage{textcomp}
\usepackage{amsmath, bm}
\usepackage{times}
\usepackage{latexsym}
\usepackage{gensymb}
\usepackage[boxed]{algorithm2e}
\usepackage{graphicx,subfigure}
\usepackage{graphicx}
\usepackage{times}
\usepackage{url}
\usepackage{latexsym}
\usepackage{amssymb}
\usepackage{amsmath}
\usepackage{multirow}
\usepackage{url}
\newcommand{\mathleft}{\@fleqntrue\@mathmargin0pt}
\usepackage{color, colortbl}
\definecolor{Gray}{gray}{0.9}
\newcolumntype{g}{>{\columncolor{Gray}}c}

\usepackage{xcolor}
\def\BibTeX{{\rm B\kern-.05em{\sc i\kern-.025em b}\kern-.08em
    T\kern-.1667em\lower.7ex\hbox{E}\kern-.125emX}}
\begin{document}

\title{TabSim: A Siamese Neural Network for Accurate \\ Estimation of  Table Similarity}

\author{\IEEEauthorblockN{Maryam Habibi}
\IEEEauthorblockA{\textit{Humboldt Universit{\"a}t zu Berlin} \\
habibima@informatik.hu-berlin.de}
\and
\IEEEauthorblockN{Johannes Starlinger}
\IEEEauthorblockA{\textit{Humboldt Universit{\"a}t zu Berlin} \\
starlinger@informatik.hu-berlin.de }
\and
\IEEEauthorblockN{Ulf Leser}
\IEEEauthorblockA{\textit{Humboldt Universit{\"a}t zu Berlin} \\
leser@informatik.hu-berlin.de}
}

\maketitle

\begin{abstract}
Tables are a popular and efficient means of presenting structured information. They are used extensively in various kinds of documents including web pages. Tables display information as a two-dimensional matrix, the semantics of which is conveyed by a mixture of structure (rows, columns), headers, caption, and content. Recent research has started to consider tables as first class objects, not just as an addendum to texts, yielding interesting results for problems like table matching, table completion, or value imputation. All of these problems inherently rely on an accurate measure for the semantic similarity of two tables. We present \textit{TabSim}, a novel method to compute table similarity scores using deep neural networks. Conceptually, \textit{TabSim} represents a table as a learned concatenation of embeddings of its caption, its content, and its structure. Given two tables in this representation, a Siamese neural network is trained to compute a score correlating with the tables' semantic similarity. To train and evaluate our method, we created a gold standard corpus consisting of 1500 table pairs extracted from biomedical articles and manually scored regarding their degree of similarity, and adopted two other corpora originally developed for a different yet similar task. Our evaluation shows that \textit{TabSim} outperforms other table similarity measures on average by app. 7\% pp F1-score in a binary similarity classification setting and by app. 1.5\% pp in a ranking scenario. 
\end{abstract}


\section{Introduction}

Digitalization facilitates management and manipulation of large-scale data sets, such as large collections of documents, audio recordings, or images. In such large collections, finding specific objects efficiently is only possible with computational tools. The predominant form of searching is based on the similarity of objects, where an algorithm would identify and rank a list of objects from the collection based on their similarity to a given query object. Similarity search requires object type specific similarity measures. For instance, some form of textual similarity may be used for searching document collections whereas a measure for time-series similarity would be employed for searching collections of audio recordings.
	
A particularly valuable type of information object are tables, which only recently have received appropriate attention. Tables are used to present structured information in a two-dimensional matrix, and are extensively used in scientific articles, business reports, product specifications, web pages etc. Research on tables as first class objects started roughly 10 years ago with the availability of large table collections, mostly extracted from web pages~\cite{gonzalez2010google, cafarella2008webtables, lehmberg2016large} or from Wikipedia~\cite{bhagavatula2013methods}. 

This work is concerned with table similarity (TS): Given a pair of tables, e.g. a query table and a table from a table corpus, compute an accurate estimate of their semantic similarity. 
TS is a fundamental operation and a prerequisite for many further applications, such as table clustering and classification~\cite{yoshida2001method, Vilain2006}, table auto-completion~\cite{agassi2004auto}, table fusion~\cite{gonzalez2010google} or filling missing values in databases~\cite{ritze2015matching}. Despite the importance of TS, it has received only little attention as an operation in its own right so far. Existing table similarity functions are all tightly integrated into their downstream application and were not compared to other TS methods. For instance, previous works on table augmentation~\cite{yakout2012infogather,zhang2013infogather+}, table union~\cite{das2012finding}, table extension~\cite{lehmberg2015mannheim,kleppmann2018density} or table imputation~\cite{ritze2015matching} all incorporate specific TS algorithms whose individual quality is unknown. Note that TS, as we define it and as necessary for such applications, is different from the related field of table-keyword similarity~\cite{ cafarella2009data, pimplikar2012answering, nguyen2015result, gao2017scientific, zhang2018ad, zhang2019table2vec}. 

What is lacking is a general and robust method to assess the similarity of two tables. Compared to similarity of other types of objects, TS has its own, specific properties. In contrast to pure texts, where the sequence of words, sentences, and paragraphs conveys meaning, tables impose meaning through the arrangement of values in columns and rows, often augmented with header information. In contrast to image similarity, where the relative positions of pixels is extremely important, table similarity often is independent of the order of rows or columns - two tables of patients from two hospitals will be considered similar irrespective of the order in which patients appear as rows, or the order in which metadata of the patients is recorded.

In this paper, we present \textit{TabSim}, a TS method which employs deep learning techniques to achieve two main objectives: a) to generate suitable table representations, and b) to use these representations to learn an accurate similarity function for pairs of tables. It is based on Siamese neural networks, which are known to be able to learn a similarity model given only few samples~\cite{koch2015siamese}. \textit{TabSim} does not require any hand-crafted features, but learns a similarity function directly from a gold standard corpus. \textit{TabSim}'s network first generates a representation of each table as a concatenation of embeddings of its caption and of its tabular content. For these, we apply two different networks to properly reflect their diverging structures:
A Bidirectional LSTM (Bi-LSTM) layer capable of modeling sequences is utilized to capture semantic information from captions because the order of words in the caption carry semantic information. Tabular data is represented by an order-invariant self-attention neural network, since the order of cells within a column and the order of columns within a table most often does not carry meaning. The two representations are shared by both compared tables to guarantee the symmetry of the similarity score. Model parameters are optimized with a contrastive loss function that relies on tables distances.

To train and evaluate \textit{TabSim}, we created a novel corpus consisting of 1500 table pairs extracted from biomedical articles and manually scored regarding their pairwise degree of similarity. To the best of our knowledge, this is the first publicly available gold standard corpus for TS. We also evaluated our approach on two other corpora originally developed for a different yet similar task which allowed adaptation: a) tables extracted from arXiv articles and b) tables in Wikipedia pages. Our evaluation on these three corpora shows that, on average, \textit{TabSim} outperforms baselines by app. 7\% pp F1-score in a binary similarity classification setting and by app. 1.5\% pp in a ranking scenario using NDCG. 

The paper is organized as follows.  In Section 2, we review existing techniques for TS. We explain \textit{TabSim} and its neural architecture in Section 3. Section 4 presents data preparation, the used baselines, and evaluation settings and metrics. In Section 5, we provide the results of our evaluation and conclude in Section 6. 

\section{Related Work}
The problem of table similarity is mostly researched in two contexts: a) table keyword search where the query is a set of keywords whose similarity to tables in a corpus must be computed, and b) table similarity search where also the query is a table. Most previous studies focused on the former. 

\subsection{Table Keyword Search}
Current table keyword search techniques mostly address a particular class of tables called relational tables, where each row corresponds to an entity and each column represents a distinct attribute of the entities. 
 Most techniques for keyword search in tables apply a two-phase approach, where a first phase of searching the textual contents of tables using a keyword query is followed by a second phase of filtering and ranking. Earlier techniques used keyword-based search and subsequent filtering of results based on schema matching techniques~\cite{cafarella2008webtables, cafarella2009data,pimplikar2012answering,nguyen2015result}. Gao et al. ~\cite{gao2017scientific} introduced a probabilistic ranking framework and Zhang et al.~\cite{zhang2018ad} used word embeddings and graph embeddings to represent tables and a learning-to-rank algorithm to re-rank initial retrieval results. Recently, Zhang et al.~\cite{zhang2019table2vec} proposed table embeddings (similar to word embeddings but with a different notion of the context surrounding each token) to represent tables, showing that such embeddings can improve search results. Another line of research tries to improve table keyword search by assigning different priorities to different table units (metadata, caption, headers, etc).
 These techniques typically assign weights to each unit using either heuristics~\cite{pyreddy1997tintin} or frequency-based methods~\cite{liu2007tablerank}.

\subsection{Table Similarity Search }
All methods for table similarity search we are aware of were developed for entity tables and are tightly bound to specific downstream applications, such as table extension, table imputation, table augmentation, table union, or table summarization. Most of the published methods use variations of Jaccard's similarity index to measure the similarity of a pair of tables. For instance, Lehmberg et al.~\cite{lehmberg2015mannheim} apply Jaccard's similarity to the tables' contents. Work on table extension~\cite{kleppmann2018density} and table imputation~\cite{ritze2015matching} uses a Jaccard similarity measure that depends on the identification and intersection of entities from DBPedia~\cite{auer2007dbpedia} in tables. Other methods were designed to build a table graph for table augmentation and information extraction~\cite{yakout2012infogather,zhang2013infogather+}. Here, similarity of two tables is computed by first applying schema matching and then computing the ratio of the tables' common attributes using cosine similarity. Another variation was presented for computing table unions. Here, similarity is computed based on Jaccard similarity of columns followed by an aggregation of similarity scores of pairs of columns using a maximum weight matching algorithm~\cite{das2012finding}. In another work, a TS was proposed for the table selection and summarization task with explicit account of both schema similarity and diversity~\cite{nguyen2015result}. 

Recently, a specific deep learning model has been developed to discern table relations like equivalence or subset~\cite{fetahu2019tablenet}. They represent table columns by a Bi-LSTM layer and identify the relation between the columns of two tables using attention layers and a softmax classifier. However, this method does not result in a symmetric similarity function, as the weights of the candidate table's Bi-LSTM layer are initialized by the output of 
the Bi-LSTM layer of the query table.

In contrast to previous work, the \textit{TabSim} TS method presented in the following automatically learns table representations and table similarity functions from training samples. It does not rely on any extra (and potentially erroneous) pre-processing steps like schema matching or entity identification and entity linking. It builds on the idea of table embeddings, but jointly computes embeddings for pairs of tables, not individual tables, to fully exploit the similarity information in the training data. And unlike \cite{fetahu2019tablenet}, \textit{TabSim} is a symmetric measure, making it applicable to a wide range of downstream applications.

\section{TabSim}
The problem of table similarity can be modeled as a one-shot classification problem~\cite{fei2006one} commonly used for face recognition, where, given only a small number of training samples, the learned similarity model must be capable of accurately classifying unseen samples. We propose to use a one-shot learning framework to learn a TS measure given low numbers of training pairs. To this end, we harness the power of Siamese deep neural networks~\cite{bromley1993signature} to extract relevant table features and learn the similarity metric. Given two tables, the network produces a distance score which is equal or larger than zero. The score zero denotes full similarity, while larger scores indicate increasingly smaller similarity (and increasing distance). These scores can also be used for classification, where pairs with a score below a given threshold are treated as similar and all others as dissimilar.

In \textit{TabSim}, each table is represented by the concatenation of two neural networks, one extracting tabular content $t$ and the other one modeling table captions $c$. The weights within these networks are shared by two tables to ensure that the similarity model is symmetric. The computed distance of two table representations is fed to a contrastive loss function, which ensures that pairs of semantically similar tables are placed in a close distance. In the following, we describe the shared layers for representation of tabular content and captions, and the Siamese neural network with the contrastive loss function used for similarity learning.

\subsection{Caption Representation}
Each table caption is modeled as a fixed-sized (cropped or padded by zero) one dimensional array, consisting of $|T_{c}|$ tokens.
The tokens of the caption are first represented by an embedding vector to map word tokens into a lower-dimensional space capturing the frequencies of co-occurring adjacent words. These are passed to a Bidirectional Long Short-Term Memory (LSTM~\cite{hochreiter1997long}) layer (Bi-LSTM~\cite{schuster1997bidirectional}) to model the relation between tokens in the caption.
 We describe the embedding layer and the Bi-LSTM layer in detail in the following:

\subsubsection{Embedding Layer}

Each caption $c$ is first represented by the sequence of tokens $T_{c}=\{w_1,\cdots,w_{|T_c|}\}$. Each token $w_t$ is represented by a binary vector with length $|V|$ (vocabulary size), where all the elements of the vector are set to zero except the index of the token in the dictionary which is set to one.
An embedding layer,
 maps each of these binary vectors into a lower dimensional vector space $e_t=W_ew_t$,
where $W_e$ is the mapping weight matrix and $e_t$ is the representation of token $w_t$ in the embedding space. The weight matrix $W_e$ is initialized with a pretrained word embedding model (See Section~\ref{subsec:es}) and is retrained in our network.

\subsubsection{Bi-LSTM Layer}
The output of the embedding layer is a sequence of tokens in embedding space. Each sequence is sent into a Bi-LSTM network to model the semantic dependencies between the tokens in the sequence from both directions. The Bi-LSTM layer consists of two LSTM layers, one in forward direction and one in backward direction. 
%
%
%
The output of the Bi-LSTM layer is designated as a representation $\phi(c)=h_{|T_c|} \oplus h'_{|T_c|}$ for caption $c$, where $h_{|T_c|}$ and $h'_{|T_c|}$ are the last output of the forward
and backward LSTM layers.

\subsection{Tabular Content Representation}
\label{subsec:TCR}
A tabular input contains both header and data cells.
In the following, we assume a horizontal table layout, where all table headers are located in the first row. Note that \textit{TabSim} can equally be applied to tables with vertical layout with headers in the first column by first identifying the tables' orientations using a current technique~\cite{eberius2015building,nishida2017understanding,crestan2011web}, and transforming them to horizontal layout by rotation. 
We also presume that all columns have the same number of cells. However, the method is also applicable to tables where subsequent rows in some of the table's columns are merged. This is done by substituting each merged cell with $s$ cells in vertical direction with the same content where $s$ is the span size of the merged cell~\cite{zhang2018ad}.
The tabular input, here, is modeled as a fixed-sized (truncated or padded by zero) three dimensional array, consisting of $N$ rows, $M$ columns and $|T_{u}|$ tokens in each cell. 

\begin{figure}[th]
\centering
\includegraphics[scale=0.46]{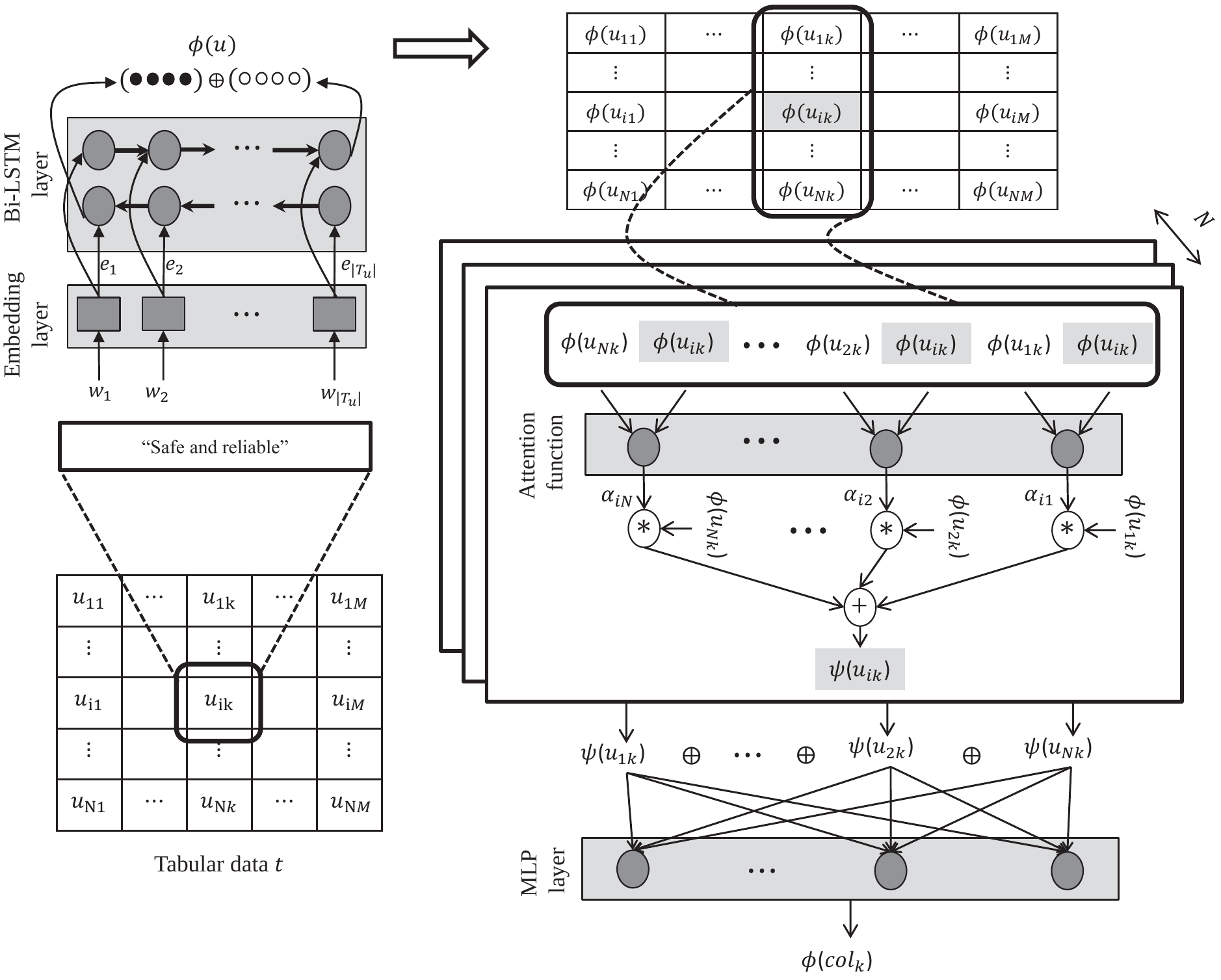}
\caption{Column representation architecture. Each cell in the tabular input $\boldsymbol{t}$ is tokenized. The tokens  of the example cell content ``safe and reliable''  in row $\boldsymbol{i}$ and column $\boldsymbol{k}$ are passed to an embedding layer and the resulting vectors are fed to a Bi-LSTM layer. The resulting output builds the cell representation $\boldsymbol{\phi(u_{ik})}$. To model the relation of cells within a column, the cell vectors within column $\boldsymbol{col_k}$ are given to an attention layer to generate weights for all the cells in the column with respect to one of the cells. The new cell representation $\boldsymbol{\psi(u_{ik})}$ is calculated by the weighted sum of all  cell vectors in the column. The new representations are concatenated and given to an MLP neural network. The resulting output is the column representation $\boldsymbol{\phi(col_k)}$.}
\label{fig:rc}
\end{figure}

A cell contains a sequence of word tokens where the order of words carries meaning. To infer this meaning, the tokens of each cell $u$ are first represented by an embedding layer to map each token to a lower dimensional semantic space, and then the embedding vectors are passed to a Bi-LSTM capturing the dependencies between tokens within a cell. The resulting vector is the representation $\phi(u)$ for cell $u$ as shown on the left side of Figure~\ref{fig:rc}.

A column is made of a set of cells where the permutation of the cells within the column does not change the column's meaning. 
Recently, permutation invariant neural networks have been proposed to model the relation between the elements of a set using different aggregation schemes over the elements, such as average and maximum operations~\cite{zaheer2017deep} or self-attention layers~\cite{lee2019set}.
In \textit{TabSim}, each column is represented by a permutation invariant network shared by all columns.
This layer first represents each cell based on its context in the corresponding column. Then the cell representations are concatenated and compressed to a lower dimensional space as the column representation. This network is capable of extracting the relation between columns in a tabular structure, as all columns share a single copy of this layer.
Similarly, the permutation among the columns in a table does not change the meaning of a tabular structure. Therefore a similar permutation invariant network is utilized to aggregate column representations by taking into account the relation between columns. The resulting vector is regarded as the representation for the tabular input.
Column representation and aggregation networks are defined in detail in the following steps.

\paragraph{Column representation}

All cells within a column $col_k$ are given to a neural network to capture cell semantics as shown in Figure~\ref{fig:rc}.
As the order of cells within a column often does not carry information, an order invariant neural network is
designed to model cell relations by inspiration from set transformer architecture~\cite{lee2019set} where a permutation invariant network is proposed for set representations using self-attention neural networks.
 Similarly, we use a self-attention layer to recalculate a new representation for each cell corresponding to the attention that each cell pays to the other cells within a column (see Section~\ref{subsec:SAL}),
because self-attention neural networks ~\cite{vaswani2017attention} ignoring positional information are of permutation invariant property.
As shown on the right side of Figure~\ref{fig:rc}, given the cell $u_{ik}$ from column $col_{k}$, the new representation $\psi(u_{ik})$ with cell relation information is defined as the weighted sum of all other cell vectors, $\forall \phi(u)$ in $col_k$. The corresponding weights are obtained by an attention function.
The column representation $\phi (col_k)$ for column $col_k$ is estimated by concatenating all cell representations $\forall \psi(u)$ and passing them to a Multilayer Perceptron~\cite{hornik1989multilayer} (MLP) for dimensionality reduction (see Section~\ref{subsec:mlpl}).

\paragraph{Column aggregation}
A similar architecture is employed for column aggregation but on columns within the table instead of cells within a column.
As the order of columns within a table often does not carry meaning, another self-attention layer is applied on columns in a table $t$ to generate a new column representation $\psi (col_k)$. Then the outputs of the self-attention layer for each column are concatenated and given to a MLP layer as a tabular representation $\phi(t)$.  
In the following, we only describe the self-attention layers and the MLP layers used for column representation. 

\subsubsection{Self-attention Layer}
\label{subsec:SAL}
Self-attention neural networks can extract relevant information from cells  in a column by allowing them to attend to themselves.
Given a cell, the network calculates a weight for each cell in the column with respect to this cell.
 These weights determine the importance of other cells when representing the respective given cell in a column.
Here, the weight is calculated using an attention function and then normalized through a softmax layer.
For each cell, the obtained normalized weights are then multiplied by their corresponding cell vectors and the resultant vectors are summed up as the output of the self-attention layer for this cell. Mathematically, the new representation obtained by the self-attention network for cell $u_{ik}$ in column $col_k$ is written in Equation~\ref{eq:2}:
\begin{align} \label{eq:2}
& a_{ij} = \sigma(\phi(u_{ik})^T W_a \phi(u_{jk})+b_a) \nonumber \\
& \alpha_{ij} =\frac{\exp(a_{ij})}{\Sigma^{N}_{n=1} \exp(a_{in})} \nonumber \\
& \psi(u_{ik}) = \Sigma_{j}\alpha_{ij}\phi (u_{jk})
\end{align}

\noindent
where $\phi(u_{ik})$ and $\phi (u_{jk})$ are the cell representations of cells $u_{ik}$ and $u_{jk}$ in column $col_k$, $a_{ij}$ is the weight of cell $u_{jk}$ with respect to cell $u_{ik}$ obtained by Luong's multiplicative attention function~\cite{luong2015effective}, $\alpha_{ij}$ is the weight normalized over all weights measured with respect to cell $u_{ik}$ and $\psi(u_{ik})$ is the new cell representation obtained by the self-attention layer. $W_a$ is the attention weight matrix, $b_a$ is the attention bias, $N$ is the row size 
 and $\sigma$ stands for a sigmoid function.

\subsubsection{MLP Layer}
\label{subsec:mlpl}
The new cell representations $\forall \psi (u): u \in col_k$ are concatenated and given to an MLP layer with a single hidden layer for dimensionality reduction.
This layer combines the outputs of the self-attention layers to extract the non-linear relation between the cell representations. Moreover, the shared network extracts the relation between all columns within the table, as all columns share a single copy of the MLP layer.
The output of this layer represents a column vector. The MLP layer is defined in Equation~\ref{eq:3}.
\begin{equation}
\begin{aligned}
\phi (col_k)= ReLU(W_{f} (\psi (u_{1k}) \oplus \cdots \oplus \psi(u_{Nk}))+b_{f})
\label{eq:3}
\end{aligned}
\end{equation}

Here, $\phi(col_k)$ is the representation of column $col_k$, $\psi(u_{1k})$ and $\psi(u_{Nk})$ are the outputs of the self-attention layer for the cells in column $col_k$, $W_{f}$ and $b_{f}$ are the model parameters and ReLU stands for a rectified linear unit~\cite{glorot2011deep}. Before ReLU activation, we apply batch normalization for faster training and reducing the chance of overfitting.

\subsection{Siamese Neural Network}
\begin{figure}[h]
\centering
\includegraphics[scale=0.35]{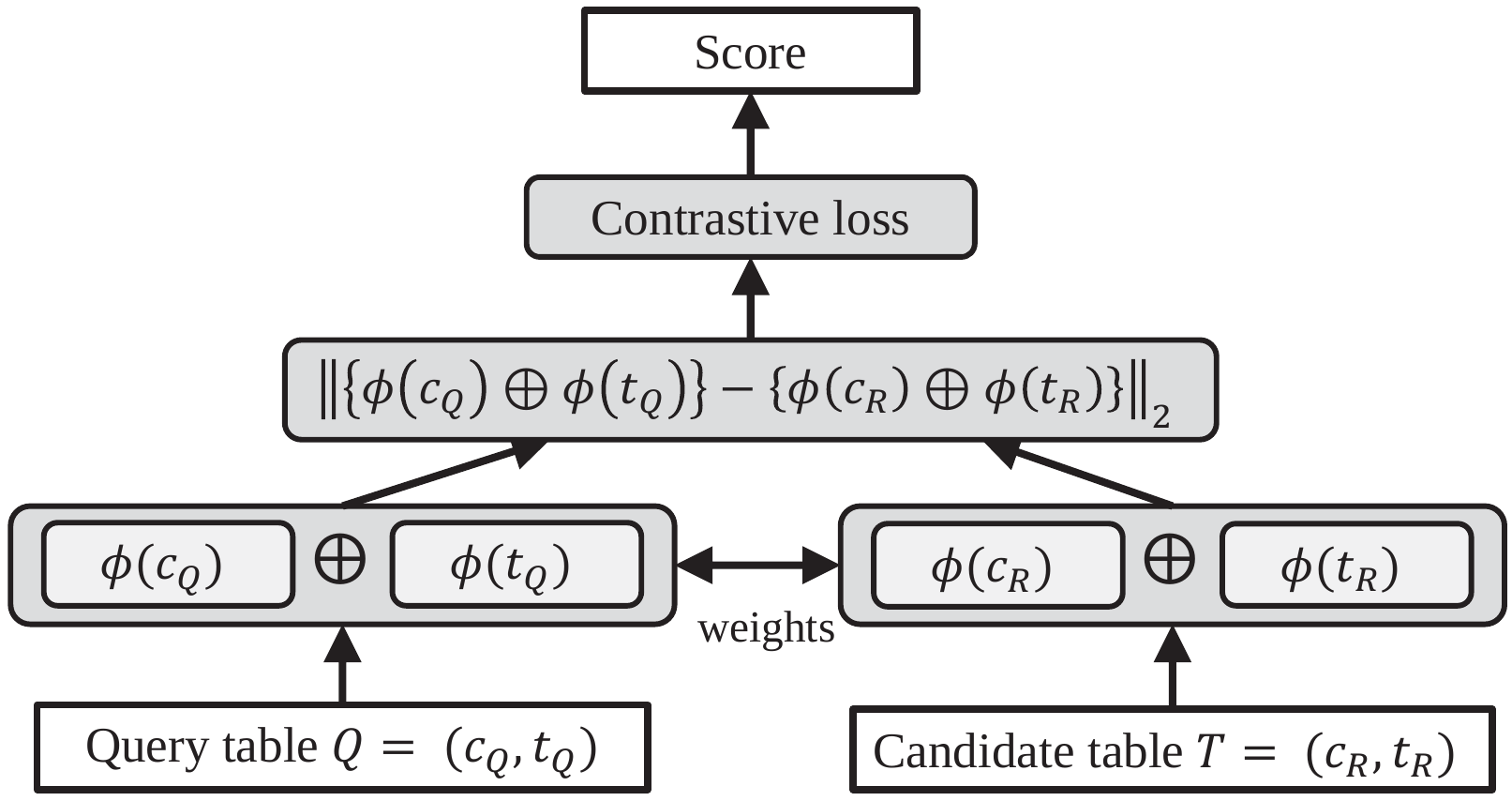}
\caption{\textit{TabSim} architecture. Caption representation $\boldsymbol{\phi(c)}$ and tabular representation $\boldsymbol{\phi(t)}$ are concatenated for tables $\boldsymbol{Q}$ and $\boldsymbol{R}$ by a network shared by both tables to ensure the symmetry of the similarity function. The distance of the resulting representations is calculated. Then this distance using a contrastive loss is minimized for similar tables and maximized for dissimilar ones.}
\label{fig:SN}
\end{figure}

Siamese  neural  networks were originally  introduced  by  Bromley et al.~\cite{bromley1994signature}  for signature verification. Given two representations, Siamese networks learn a model such that the symmetric similarity metric is small if two representations belong to the same category, and large if they belong to different categories. Here, as shown in Figure~\ref{fig:SN}, given two tables $Q=(c_Q,t_Q)$ and $R = (c_R,t_R)$, a Siamese neural network  measures  the  semantic  relatedness of the two tables as the Euclidean distance between the tables' representations, each obtained by concatenating table caption and tabular content embeddings as defined in  Equation~\ref{eq:4}.
\begin{equation}
\begin{aligned}
D_{\phi}(Q,R)= \parallel \{\phi(c_Q)\oplus \phi(t_Q)\}-\{\phi(c_R)\oplus \phi(t_R)\}\parallel_{2}
\label{eq:4}
\end{aligned}
\end{equation}

Here, $\phi(c_Q)$ and $\phi(c_R)$ are caption representations and $\phi(t_Q)$ and $\phi(t_R)$ are tabular representations for the query and the candidate table, respectively. For training, pairs with known similarity score are fed into this network. The loss function shown in Equation~\ref{eq:5} is a contrastive loss defined as quadratic function of table pair distances~\cite{hadsell2006dimensionality} so that $D_{\phi}(Q,R)$ is small (close to zero) if table $R$ is similar to table $Q$ and equal or larger than the margin $m$ otherwise.
\begin{equation}
\small
\begin{aligned}
L(y,\phi,Q,R)= \frac{1}{2} (1-y) D_{\phi}^2(Q,R) + \frac{1}{2} y \{\max(0,m-D_{\phi}(Q,R))\}^2
\label{eq:5}
\end{aligned}
\end{equation}
Here, $y$ is the true label of the pair. Labels for a binary classification are obtained by thresholding the distance at half of the margin, $\frac{m}{2}$, where those pairs with distance scores below this threshold are considered as similar, the others as dissimilar.

\section{Data and Evaluation Setup}
In this section, we describe the corpora, the evaluation metrics, the competitor methods, and the model parameter settings we used for evaluating the effectiveness of \textit{TabSim}.

\subsection{Data}
\label{subsec:data}
We performed our evaluation on three corpora. One is a new gold standard corpus for estimating table similarity we created from tables in biomedical articles. The two other corpora originally were developed for different yet similar tasks and were adapted to suit our purpose of evaluating table similarity functions.

\paragraph{PMC Gold standard table corpus}
We built, to the best of our knowledge, the first dedicated corpus for evaluating table similarity methods. Tables for our corpus were extracted from scientific articles published in the PubMed Central (PMC) Open Access subset\footnote{See https://www.ncbi.nlm.nih.gov/pmc/tools/openftlist/} which currently consists of approximately 1.5 million full-text articles from several domains in the area of biomedicine and life sciences. We randomly chose 150 query tables and manually compared each to 10 candidate tables that were obtained using the state-of-the-art table similarity algorithm from \cite{liu2007tablerank}, which applies cosine similarity over the concatenation of tf$\cdot$idf vectors from caption, header, tabular data, and the paragraph in the text referring to the table. For each query table, we selected the 5 most similar and the 5 most dissimilar tables for manual annotation. Our corpus thus initially contains 1500 pairs.

To annotate the pairs, we developed an annotation guideline for instructing five annotators familiar with biomedical  research papers. Given a table pair, the tables' captions and tabular data should
be compared and labeled separately. Each label should be assigned from one of the four classes of ``2: highly similar'', ``1: similar'', ``0: dissimilar'', and ``-1: I do not know''. The label ``highly similar'' was to be assigned if the compared elements have more than 70\% similar content. A ``similar'' label should been chosen when the elements have at least 30\% similar and at least 30\% dissimilar content. If the elements have less than 30\% similar content they should be labeled as ``dissimilar''. The ``I do not know'' label was to be selected in cases where an annotator was unsure about her judgment.  The inter-annotator agreement (Fleiss Kappa~\cite{fleiss1971measuring}) averaged over caption and tabular data judgments was 0.89 measured on 20 pairs. 

For evaluation and training, we ignore all tables with at least one ``I do not know'' label in our experiments, leading to 1391 annotated table pairs. We perform two types of evaluation: One in a classification setting, where methods must assign each table pair to the labels ``similar'' or ``dissimilar'', and a ranking setting, where methods must compute ranks for a set of candidate tables given a query table. For the first setting, we assign each table pair a binary label by aggregating the threshold-determined labels of pairs of caption and tabular content. A table pair label is labeled ``dissimilar'' if both captions and tabular parts are labeled ``dissimilar'', otherwise the table pair is labeled as similar. Using this categorization, our corpus contains 542 similar and 849 dissimilar pairs.
For the second setting, we used the query tables from the corpus creation as queries and ranked the candidates by the sum of caption and table content similarity scores.

\paragraph{arXiv table corpus:} We adapted the table corpus used for table keyword search~\cite{gao2017scientific}. The tables in the corpus were extracted from arXiv articles\footnote{https://arxiv.org/help/bulk$\_$data$\_$s3} in the domain of physics. In this corpus, given a keyword query, tables are assigned into one of the four classes: ``3:highly similar'', ``2:similar'', ``1:less similar'', or  ``0:dissimilar''. The label ``highly similar'' means that the table entirely covers the information need expressed by a keyword query. A ``similar'' label means that the table provides comprehensive information on the query's topic, while ``less similar'' indicates that the table provides some information for the query topic. If the table does not provide useful information on the topic, it is labeled as ``dissimilar''.

To adapt this corpus to table
similarity classification, we iterate over all queries $q$ and consider
all tables that are highly similar or similar to $q$ to also be similar
pair-wise and dissimilar to all other tables that were compared to $q$.
This results in 836 similar and 836 dissimilar pairs. For ranking, we follow the same idea and rank pairs of tables for a query $q$ according to their similarity to $q$.

\paragraph{Wikipedia table corpus:} We adjusted the table corpus created for table alignment~\cite{fetahu2019tablenet}. The tables are extracted from Wikipedia articles\footnote{https://en.wikipedia.org/wiki/Wikipedia:Database$\_$download} without any domain focus. In this corpus, pairs of tables are assigned to one of the three classes ``2:equivalent'', ``1:subPartOf'', or ``0:noalignment''. The label ``equivalent'' means both table schemas have semantically similar columns. The ``subPartOf'' label shows that the schema of one table is a subset of the schema of the other one. If there is no relation between the schemas of the two tables, the pair is labeled ``noalignment''. For classification, we assume  all pairs with ``equivalent'' or ``subPartOf'' label as similar and the remaining as dissimilar, yielding 8162 similar and 8882 dissimilar pairs. In the relation ranking assessment scenario, the rank values are 2, 1 and 0 for ``equivalent'', ``subPartOf'' and ``noalignment'' labels respectively.

\subsection{Evaluation  Metrics}
\label{subsec:EM}

We perform the evaluation in two ways. First, we assess the performance of the models as a binary classifier (classification-based evaluation, CBE). To this end, we convert similarity scores to a binary label as described with the different methods (see Section 3.3 and next section). We report  precision, recall, F1-score, and accuracy and perform area under the curve (AUC) receiver operating characteristic (ROC)  analysis by varying classification threshold. Second, we evaluate similarity scores directly in a ranking scenario (rank-based evaluation, RBE). As described in the previous section, all pairs from the corpora we used can be grouped into sets made of a query table and some candidate tables. For each such set, we rank candidate tables based on their similarity score to the query table and compare the ranks to the gold standard ranks. For evaluation, we report normalized discounted cumulative gain (NDCG). Both evaluations are performed using 5-fold cross validation (5F-CV) on the respective corpus. Note that we did not perform any hyperparameter tuning but used default values instead (see Section~\ref{subsec:es}). This clearly leaves room for further optimizations.

\subsection{Competitors}
\label{subsec:smc}
We compare \textit{TabSim} with three competitors and two baselines. In the following, we describe how we adapted each competitor to make their results comparable to \textit{TabSim}, as \textit{TabSim} is the only method that does not make any assumptions on table content. For CBE, we consider table pairs with a similarity score higher than a threshold as similar and otherwise as dissimilar, applied to all methods. 

The first method was proposed by Lehmberg~et~al.~\cite{lehmberg2015mannheim}  for finding table extensions that includes a table similarity function based on the Jaccard similarity between entities within tables. For comparison, we adapted the method, keeping the same similarity function but representing a table as bag of words of its cells instead of only entities. Caption similarity is measured as Jaccard similarity among the bag-of-words representations of the captions. The two scores (caption and tabular content) are averaged to determine the tables' similarity score (denoted \textit{Jaccard}). 

The second method is based on the Google Fusion Tables project~\cite{das2012finding}, which determines table similarity as the normalized maximum weight matching score between the two table's columns. Weights are measured by a variant of Jaccard similarity between entities of every column pair, where a column is represented by aggregating sets of labels obtained from databases for each entity. We adapted the algorithm by representing table columns by the sum over the embeddings of all words in the column, keeping the table matching untouched. Caption similarity is computed as cosine similarity of their summarized word embeddings. The two scores are averaged as the overall score (denoted \textit{Google Fusion}).

The third approach was proposed by Liu~et~al.~\cite{liu2007tablerank}, which measures table similarity by the cosine similarity function over tf$\cdot$idf vectors of the table's content. As for \textit{Google Fusion}, we only changed the representation of tables by using the average of embedding vectors instead of tf$\cdot$idf vectors to enhance the semantic sensitivity of the method. Caption similarity is also based on embeddings. The overall similarity score is averaged over caption and tabular similarities (denoted \textit{Cosine}). 

We also compare \textit{TabSim} to two baseline models using more traditional supervised machine learning techniques. For these, we initially represent each query and candidate table using separate feature vectors for captions and content. Each of these is formed as concatenation of the respective tf$\cdot$idf vectors and word embedding vectors. We compute the absolute difference of these vectors for each pair, i.e., content minus content vector and caption minus caption vector, and pass the concatenation of the difference vectors to a supervised classifier for classification. We report results for using (a) Logistic Regression (denoted \textit{LR})~\cite{rubinstein1997discriminative} and (b) Random Forests (denoted \textit{RF})~\cite{ho1995random}. For \textit{LR}, we use the classification result for the CBE and the probability scores for RBE. For \textit{RF}, recall that Random Forests generate a probability score for each class label. We use the average of these scores over all trees (here, it contains 100 trees) for RBE, and the \textit{RF} classification result for CBE.

\subsection{Implementation}
\label{subsec:es}

We use the Keras tokenizer\footnote{https://keras.io/preprocessing/text/} for cell content tokenization.
We also implement \textit{TabSim} using Keras in Python with a Tensorflow backend. The network is trained using Keras RMSprop optimizer with default parameters.  We set the number of rows and columns in \textit{TabSim} to $N=M=9$, the number of tokens in a cell to $|T_{u}|=4$, and  the number of tokens in a caption to $|T_{c}|=12$, as, on average, the tables in our new PMC gold standard corpus have 9 rows, 10 columns, 2 tokens per cell and 12 tokens per caption. Because table sizes (and average table sizes over a given corpus) vary, we study the impact of these parameters separately in Section~\ref{subsec:ETS}. The fixed-sized table assumption allows to take advantage of vectorization operations in our implementation. We choose a default value of 1 as the margin $m$~\cite{nicosia2017accurate}. The size of the embedding layer is fixed to 200 (see below). We set the size of the Bi-LSTM hidden layer and the output of the MLP layers to 100. 
\textit{LR} and \textit{RF} are implemented using scikit-learn\footnote{https://scikit-learn.org/stable/} using default hyperparameter settings. 

We use the embeddings trained on a combination of PubMed abstracts\footnote{See https://www.ncbi.nlm.nih.gov/pubmed/} (nearly 23 million biomedical abstracts), PMC articles (nearly 700,000 full texts from the biomedical domain) and  approximately four million English Wikipedia articles\footnote{See https://dumps.wikimedia.org/} in the general domain to have a broad coverage of the topics in all three evaluation corpora. These embeddings are used by the baselines for caption and tabular representations and by \textit{TabSim} for the initialization of the embedding layer. The vectors are provided by Pyysalo et al.~\cite{pyysalo2013distributional}, with 200 dimensions, and were trained by the algorithm from Mikolov et al.~\cite{mikolov2013distributed}. 

\section{Results}
\label{sec:results}

We first compare variations of \textit{TabSim} on the PMC Gold standard corpus to study the impact of different table representations and word embeddings. Then we compare \textit{TabSim} with three competitors and two baselines on three different corpora in the CBE and the RBE setup (see Section~\ref{subsec:EM}).

\subsection{Different TabSim Configurations}
We first compare different means of column representation and layer aggregation (see Section~\ref{subsec:TCR}) to study whether order-dependent or order-invariant models should be preferred. 

In the first variant of \textit{TabSim}, inspired by the table representation developed for table relation extraction~\cite{fetahu2019tablenet}, we model both cells within a column and columns within a table as sequences, i.e., their orders are important.  For this model, we replace the attention layers of \textit{TabSim} with Bi-LSTM layers. The modified architecture uses the concatenation of the last layers of a Bi-LSTM network on cells within a column as column representation. Then the column representations are fed to another Bi-LSTM layer (again replacing the respective attention layer). The concatenation of the last layers of this Bi-LSTM network and subsequent feeding to an MLP layer with a single hidden layer using ReLU activation generates the table representation. We keep the same size of 100 for the Bi-LSTM hidden layers and MLP layers. We call this variation \textit{TabSim(L)}.

In the second variant, inspired by the table representation used for table orientation classification~\cite{nishida2017understanding}, we assume that the position of a cell within a table essentially has the same meaning as the position of pixels in an image. In this architecture, all table cells are first passed to a two-dimensional convolutional neural network~\cite{lecun1989backpropagation} (CNN). Then the CNN's outputs are flattened and fed to an MLP layer with a single hidden layer and a ReLU activation as the table representation.  We set the CNN filter size to $3\times 3$, following~\cite{nishida2017understanding}. We call this approach \textit{TabSim(C)}.

We measured 5F-CV performance in terms of accuracy, precision, recall and F1-score (macro averaged over similar and dissimilar pairs). Results are shown in Table~\ref{tab:prf}. \textit{TabSim} reaches the highest accuracy and F1-score (precision, recall) with an offset of at least 1.27\% pp and 1.38\% pp (1.07\% pp,1.59\% pp) in comparison to the two variations.

\begin{table}[ht!]
\def\arraystretch{0.95}
\centering
\caption{5F-CV macro-average precision, recall, F1-score and accuracy for \textit{TabSim(L)}, \textit{TabSim(C)} and \textit{TabSim} on the PMC Corpus.}
\label{tab:prf}
\begin{tabular}{lcccc}
\hline
\textbf{Method}&\textbf{P (\%)}&\textbf{R (\%)}&\textbf{F1 (\%)} & \textbf{Acc. (\%)}
\\
\hline
\textbf{\textit{TabSim(L)}}&90.95&	88.93&	89.92&90.66
\\
\textbf{\textit{TabSim(C)}}& 91.05	&88.86	&89.93&90.66
\\
\textbf{\textit{TabSim}}&\textbf{92.12}	&\textbf{90.52}	&\textbf{91.31} &\textbf{91.93}
\\
\hline
\end{tabular}
\end{table}	

\begin{figure*}[th]
\centering
\small
\setlength{\tabcolsep}{0.7pt}
\def\arraystretch{0.0}
\begin{tabular}{ccc}
\includegraphics[scale=0.49]{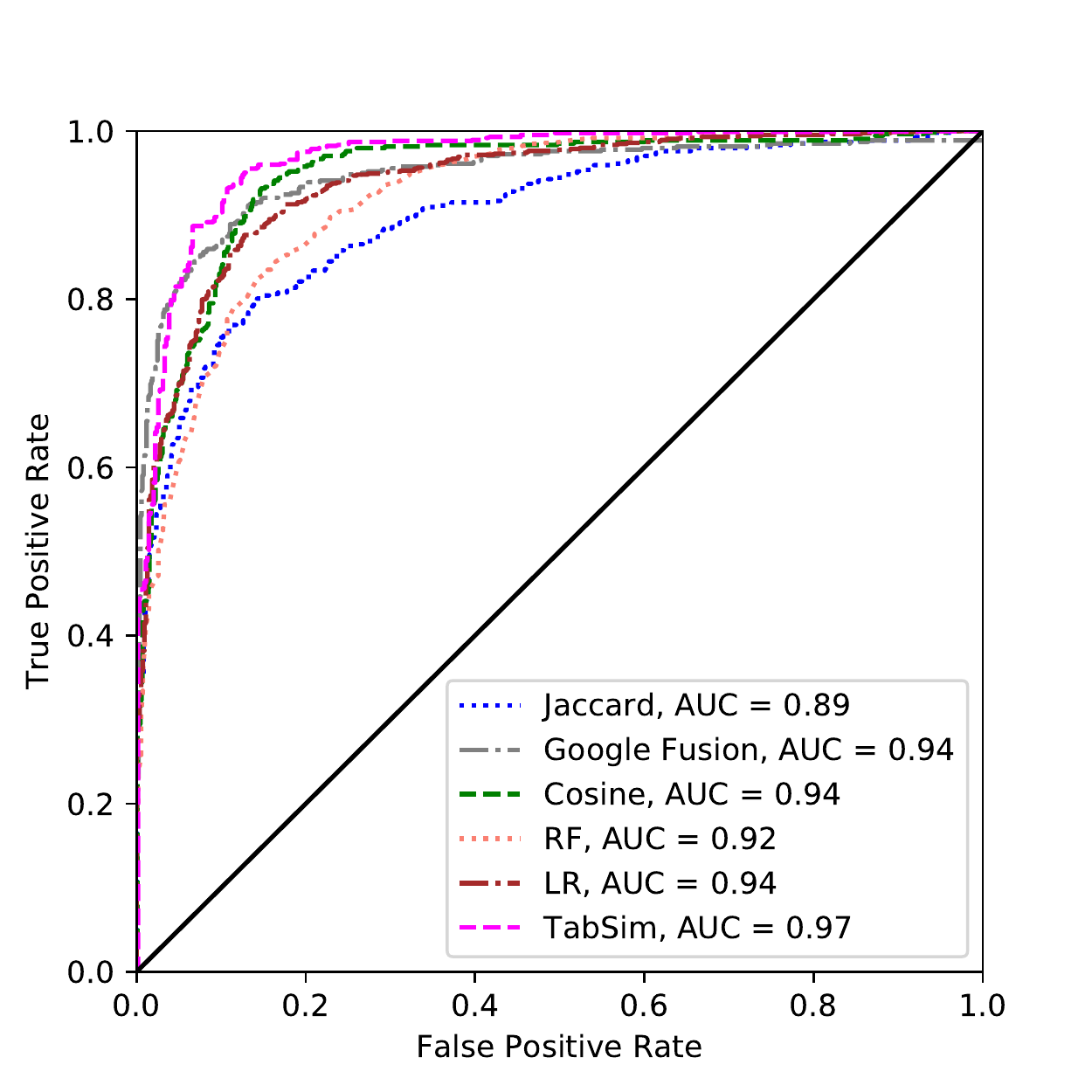}&\includegraphics[scale=0.49]{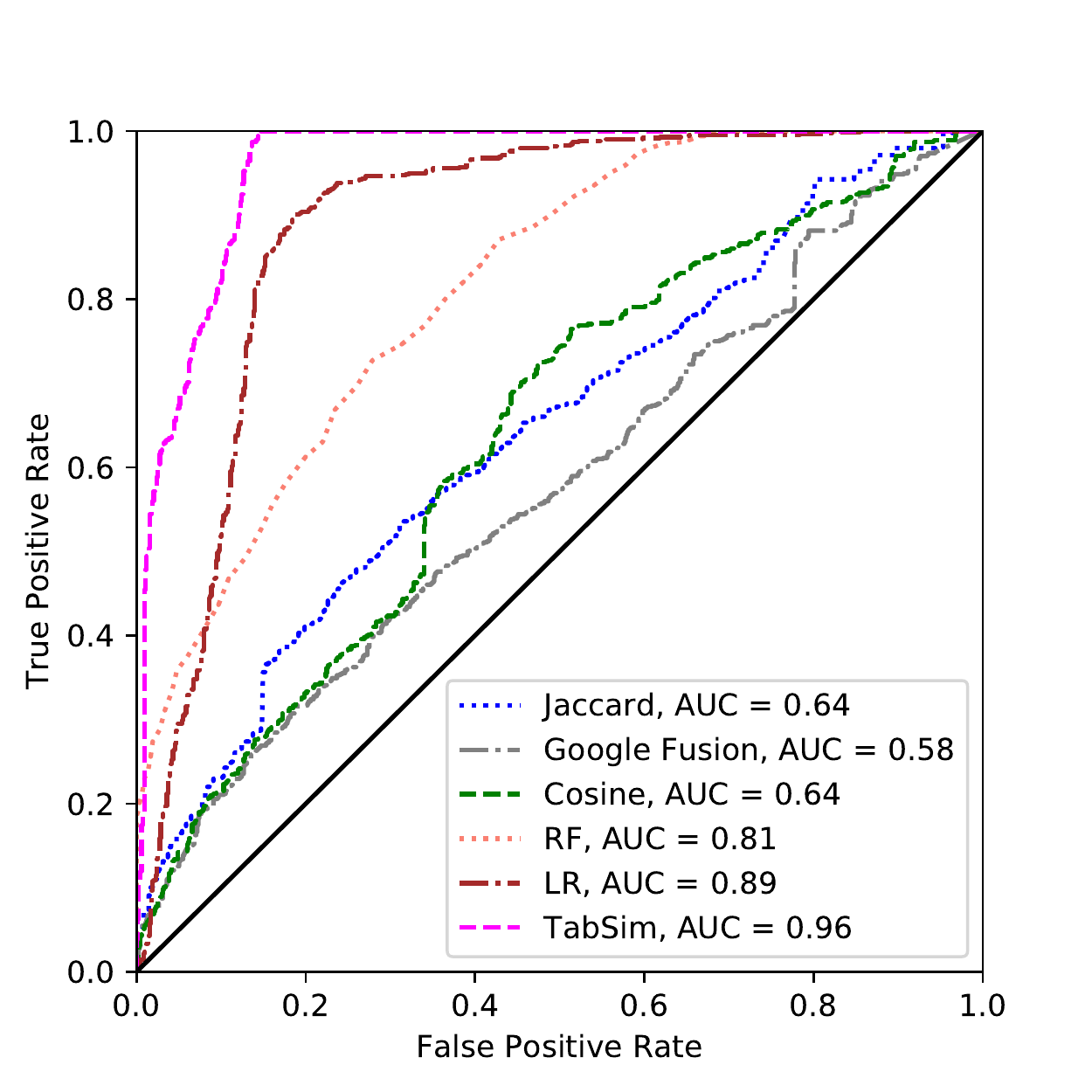}&\includegraphics[scale=0.49]{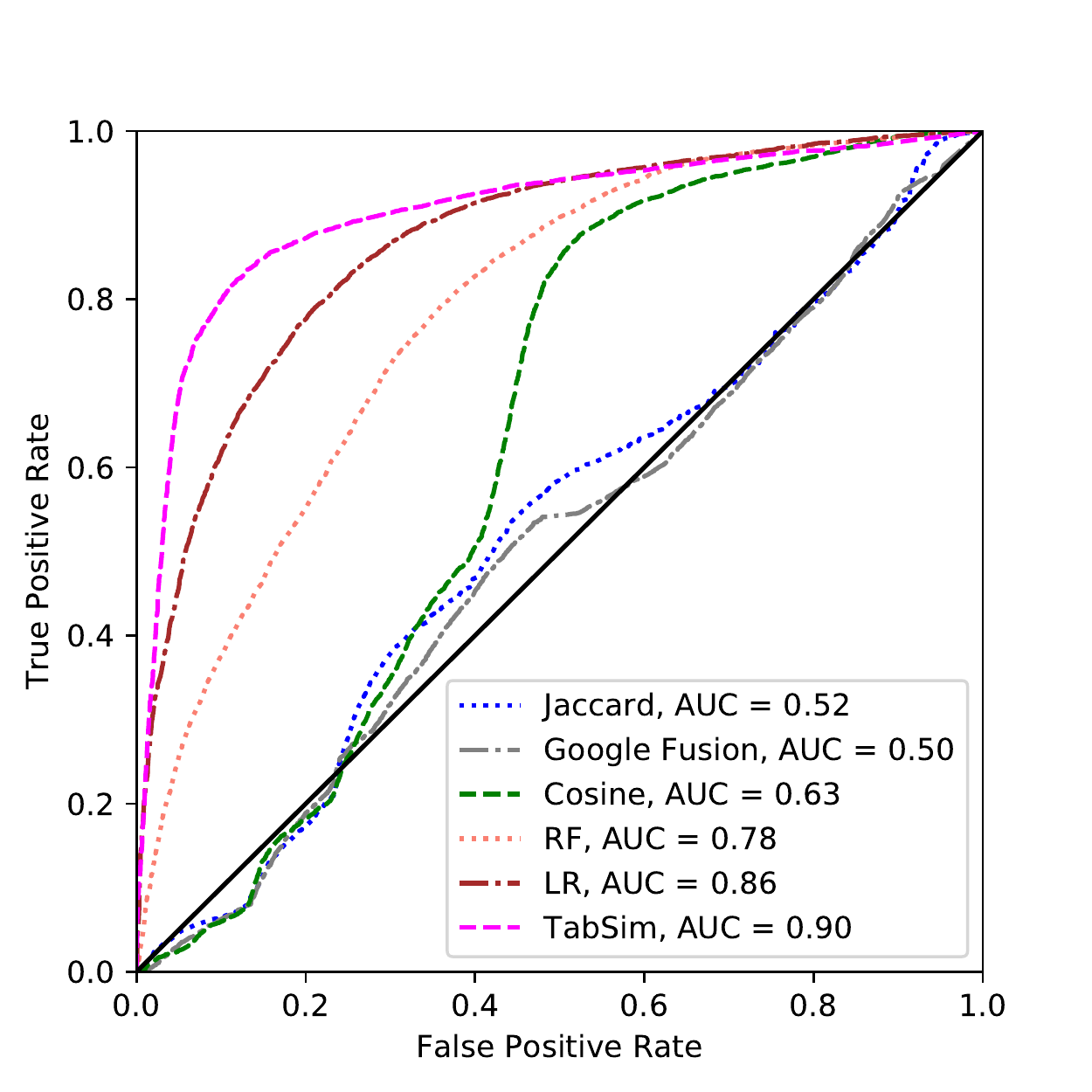}\\
(a) & (b) & (c) \\
\end{tabular}
\caption{ROC curves with AUC values for \textit{TabSim} and five other methods over (a) PMC, (b) arXiv and (c) Wikipedia.}
\label{fig:rocs}
\end{figure*}

\subsection{Impact of Pretrained Word Embeddings}

We assess the effect of differently pretrained word embeddings on the performance of \textit{TabSim} by initializing the embedding layer in four different ways: (a) random initialization, (b) word embeddings pretrained on PMC, PubMed and Wikipedia (see Section~\ref{subsec:es}), (c) word embeddings pretrained on tables and (d) word embeddings pretrained on table columns. For approach (c), word embeddings were trained on around 1.5 million tables extracted from PMC following the model proposed by Zhang et al.~\cite{zhang2019table2vec}, where word co-occurrences are modeled by word embeddings at the table level. For (d), we modify this approach by modeling word co-occurrences at the column level. Due to the order-invariant property of cells in a column (see previous section), we create different training samples by random permutation of cells within a column. We repeat this procedure 10 times per cell to increase its contextual information.

5F-CV results are shown in Table~\ref{tab:wem}. Pretrained word embeddings ameliorate F1-score (precision, recall) by at least 5.41\% pp (6.82\% pp, 4.01\% pp) and accuracy by at least 5.86\% pp compared to a random initialization. This indicates that semantic information represented by word embeddings derived from a larger corpus is essential for estimating table similarity. The use of embeddings trained on tables instead of the ones trained on articles slightly improves the scores. Embeddings that also consider the column contexts are even better, in all four metrics. 

\begin{table}[ht!]
\def\arraystretch{0.95}
\centering
\caption{5F-CV performances on the PMC Corpus using different initialization strategies for the  embedding layer: (\textit{a}) Random initialization, (\textit{b}) embeddings trained on articles, (\textit{c}) on tables, and (\textit{d}) on columns.}
\label{tab:wem}
\begin{tabular}{lcccc}
\hline
\textbf{Method}&\textbf{P (\%)}&\textbf{R (\%)}&\textbf{F1 (\%)} & \textbf{Acc. (\%)}
\\
\hline
 \textbf{a}&	85.30	&86.51	&85.90& 86.07
\\
\textbf{b}&92.12	&90.52	&91.31 &91.93
\\
 \textbf{c}&	92.64	&90.62	&91.62 & 92.18
\\
\textbf{d}& \textbf{93.01}&	\textbf{91.25}	&\textbf{92.12} & \textbf{92.66}
\\
\hline
\end{tabular}
\end{table}

\subsection{Effect of Table Size and Table Caption}
\label{subsec:ETS}
We create several versions of \textit{TabSim} by varying the number of rows and columns considered by the model. These versions are denoted \textit{TabSim}($N$,$M$) for $N=\{9,15\}$ and $M=\{9,15\}$ where  $N$ is the number of rows and $M$ is the number of columns. The chosen values for $N$ and $M$ are motivated as follows. We select 9 as one option because this is the averaged count over both rows and columns in the PMC corpus. As the second option we add to this half of the averaged standard deviation of rows and columns in PMC. We present the results from 5F-CV in Table~\ref{tab:tse}. The results show that there is no significant change in performance resulting from the truncation implied by the table size parameters. We conclude that choosing the mean values as the table size in our fixed-sized network is an appropriate selection, which also allows us to still keep the number of model parameters small.

\begin{table}[ht!]
\def\arraystretch{0.95}
\centering
\caption{5F-CV classification performance for \textit{TabSim}($\boldsymbol N$,$\boldsymbol M$) with four different sizes $\boldsymbol{N=\{9,15\}}$ and $\boldsymbol{M=\{9,15\}}$.}
\label{tab:tse}
\begin{tabular}{lcccc}
\hline

\textbf{Method}&\textbf{P (\%)}& \textbf{R (\%)}&\textbf{F1 (\%)}&\textbf{Acc. (\%)}
\\
\hline

\textbf{\textit{TabSim(9,9)}}&	\textbf{92.12}&\textbf{90.52}& \textbf{91.31}	&\textbf{91.93}	
\\
\textbf{\textit{TabSim(9,15)}}&	91.41&89.46& 90.41	&91.10	
\\
\textbf{\textit{TabSim(15,9)}}&91.67&89.79&90.71&	91.40	
\\
\textbf{\textit{TabSim(15,15)}}&91.56&89.72&	90.63	&91.29
\\
\hline
\end{tabular}
\end{table}

As a last experiment regarding the configuration of \textit{TabSim}, we created a version which ignores table captions and only relies on the table's content. This modification reduces accuracy and F1-score (precision, recall) scores by 3.68\% pp and 3.76\% pp (4.82\% pp, 2.71\% pp), which highlights the importance of table captions and table content for assessing TS.

\subsection{Binary Classification Performance}
\label{subsec:BCP}
We compare the classification performance of \textit{TabSim} with three competitors, namely \textit{Jaccard}, \textit{Cosine}, \textit{Google Fusion}, and two baselines \textit{LR} and \textit{RF}  (see Section~\ref{subsec:smc}) by measuring AUC over different classification thresholds. Figure~\ref{fig:rocs} shows ROC curves for each corpus separately, showing that \textit{TabSim} has the highest AUC in all three corpora, beating the other methods by at least 3\% pp, 7\% pp and 4\% pp on the PMC, arXiv, and Wikipedia table corpora, respectively. The three competitors \textit{Jaccard}, \textit{Cosine} and \textit{Google Fusion} are particularly worse than \textit{TabSim} on arXiv and Wikipedia. The latter two are close to \textit{TabSim} for the PMC corpus, which can be explained by the fact that this corpus was created by selecting candidate tables (for manual annotation) using cosine similarity, the similarity function which is also used by \textit{Cosine} and partly by \textit{Google Fusion}. The two baselines \textit{LR} and \textit{RF} are always outperformed by \textit{TabSim}, again with a larger margin for arXiv and Wikipedia than for PMC.

\begin{table}[ht!]
\setlength{\tabcolsep}{2pt}
\def\arraystretch{.95}
\centering
\caption{5F-CV performance at a fixed classification threshold of 0.5 for \textit{Jaccard}, \textit{Cosine}, \textit{Google Fusion}, \textit{LR}, \textit{RF}, \textit{TabSim} on three different corpora. }
\label{tab:prf}
\begin{tabular}{llcccc}
\hline
 \textbf{Corpora}&\textbf{Method}&\textbf{P (\%)}&\textbf{R (\%)}&\textbf{F1 (\%)} & \textbf{Acc. (\%)}
 \\
\hline

\multirow{6}{*}{\textbf{PMC}}& \textbf{\textit{Jaccard}}&	74.23	&72.28	&73.23	&67.35\\
&\textbf{\textit{Cosine}}&86.97&88.91	&87.93 &87.65
\\
&\textbf{\textit{Google Fusion}}&82.97	&63.17	&71.72	&70.77
\\
&\textbf{\textit{RF}}&85.46	&81.88	&83.59	&84.19
\\
&\textbf{\textit{LR}}&86.58	&87.23	&86.90	&87.43
\\
&\textbf{\textit{TabSim}}&\textbf{92.12}	&\textbf{90.52}	&\textbf{91.31}	&\textbf{91.93}
\\
\hline
\multirow{6}{*}{\textbf{arXiv}}& \textbf{\textit{Jaccard}}&	60.42	&57.25	&58.79	&57.25\\
&\textbf{\textit{Cosine}}&56.91	&56.66	&56.79	&56.66
\\
&\textbf{\textit{Google Fusion}}&55.07	&50.27	&52.56	&50.27
\\
&\textbf{\textit{RF}}&71.88	&71.80	&71.84	&71.80
\\
&\textbf{\textit{LR}}& 81.30	 &79.47	 &80.37	&79.47
\\
&\textbf{\textit{TabSim}}&\textbf{91.71}	&\textbf{91.23}	&\textbf{91.47}	&\textbf{91.23}
\\

\hline
\multirow{6}{*}{\textbf{Wikipedia}}& \textbf{\textit{Jaccard}}&60.42	&57.25	&58.79 & 52.66\\
&\textbf{\textit{Cosine}}& 50.65	&50.12	&50.38& 50.93
\\

&\textbf{\textit{Google Fusion}}& 41.80	&46.11	&43.84 & 46.84
\\
&\textbf{\textit{RF}}&70.52	&70.78	&73.01& 70.45
\\
&\textbf{\textit{LR}}	&76.97	&76.60	&76.78& 77.14
\\
&\textbf{\textit{TabSim}}&\textbf{84.24}	&\textbf{82.74}	&\textbf{83.06}& \textbf{83.60}
\\
\hline
\end{tabular}
\end{table}	

Figure \ref{fig:rocs} shows that different methods have their highest F1-measure at different thresholds. However, the optimal threshold usually is not known for corpora without gold standard annotations, in which case one has to resort to default settings. To also consider this scenario, we computed 5F-CV performance for all combinations of method/corpus using a fixed threshold 0.5; see Table~\ref{tab:prf}. Notably, \textit{TabSim} outperforms all other methods in all measures and for each corpus. Again, the differences to the competitor methods are particularly high for arXiv and Wikipedia; of these, \textit{Cosine} gets closest to \textit{TabSim} on the PMC corpus. 

One could suspect that the advantages of \textit{TabSim} (\textit{RF} and \textit{LR}) on PMC are due to the fact that they use word embeddings pretrained on a corpus that is dominated by biomedical articles, the same domain as the PMC corpus. However, this does not seem to be the case, as the advantages of \textit{TabSim} are much higher for the other two corpora than for PMC. Furthermore, the two competitors \textit{Google Fusion} and \textit{Cosine} also use the same word embeddings as \textit{TabSim}. 

\subsection{Error Analysis}
\begin{figure}[h]
\small
\setlength{\tabcolsep}{0.7pt}
\def\arraystretch{0.0}
\begin{tabular}{ccc}
\centering
\includegraphics[scale=0.37]{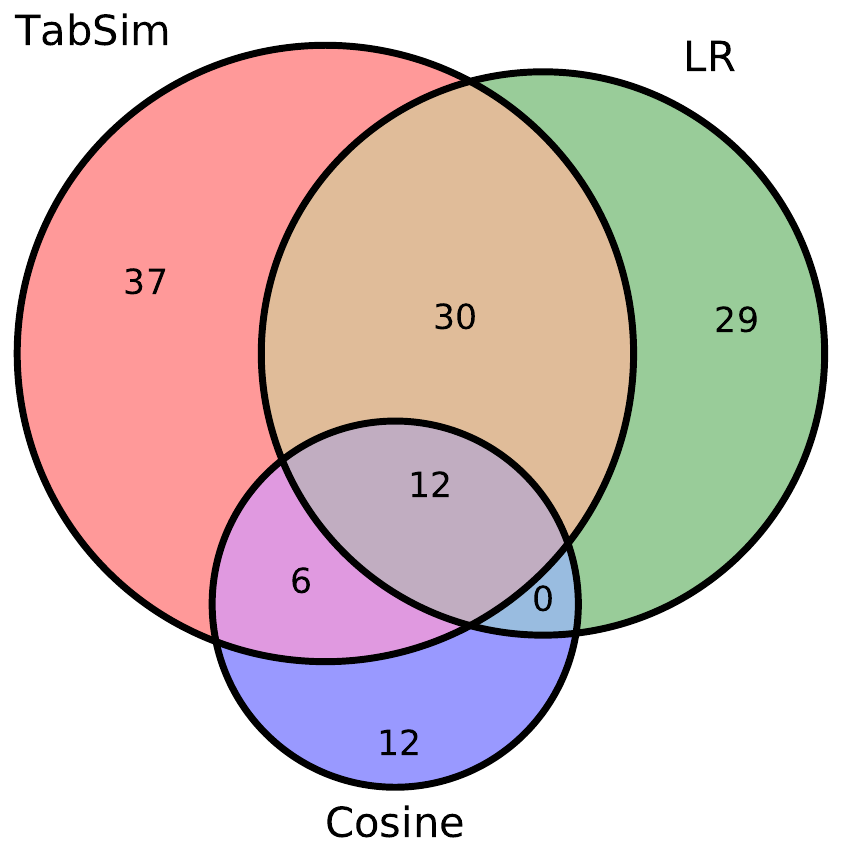} &\includegraphics[scale=0.37]{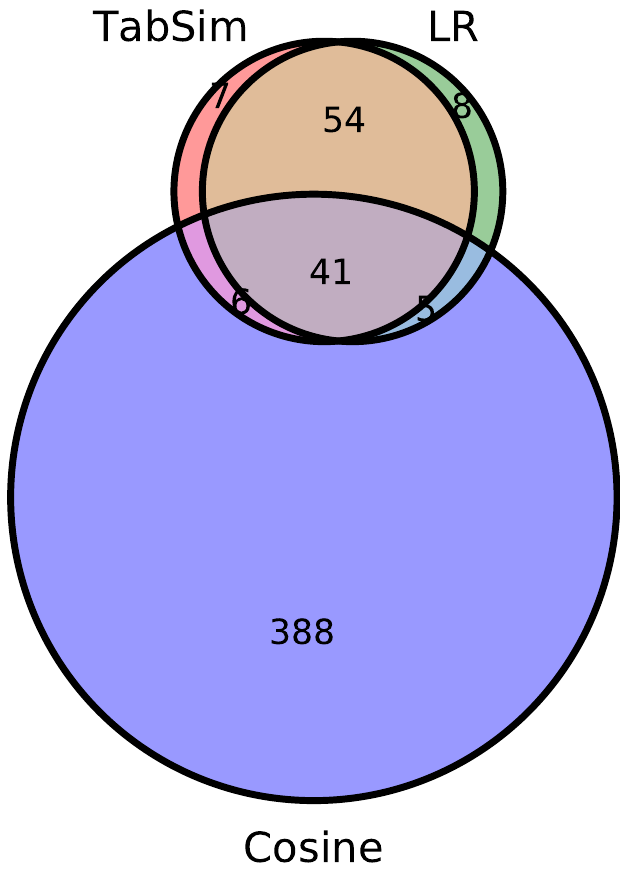}&
\includegraphics[scale=0.37]{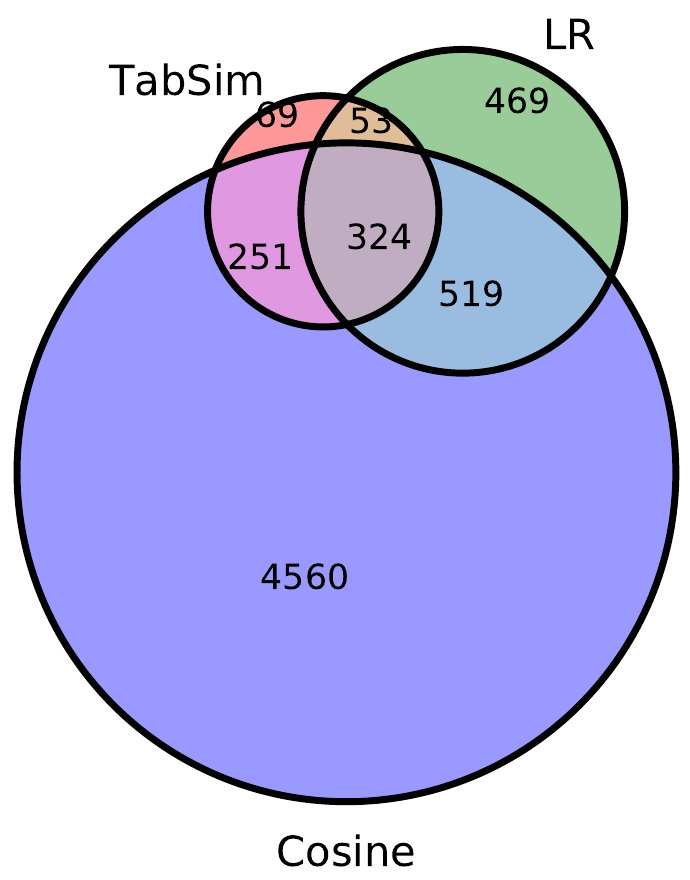}
\\
(a)&(b) &(c)\\
\includegraphics[scale=0.37]{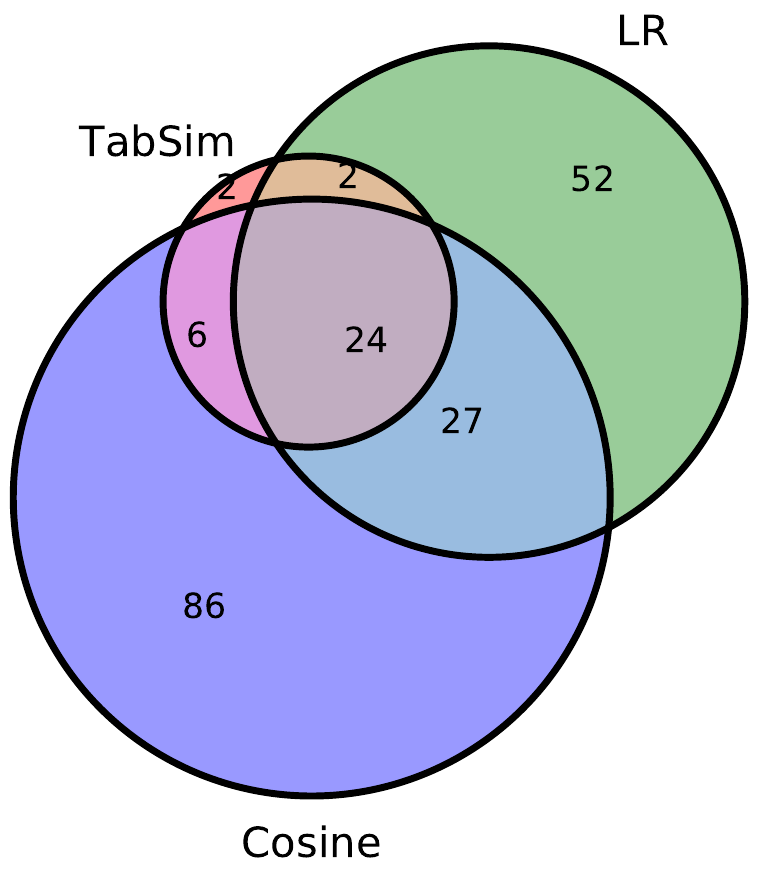} &\includegraphics[scale=0.37]{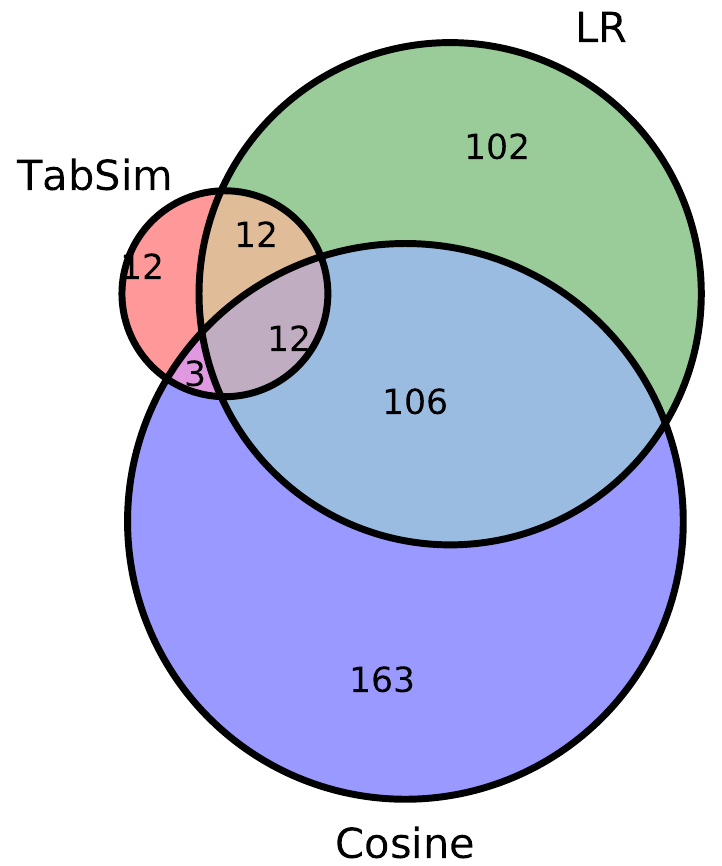}&
\includegraphics[scale=0.37]{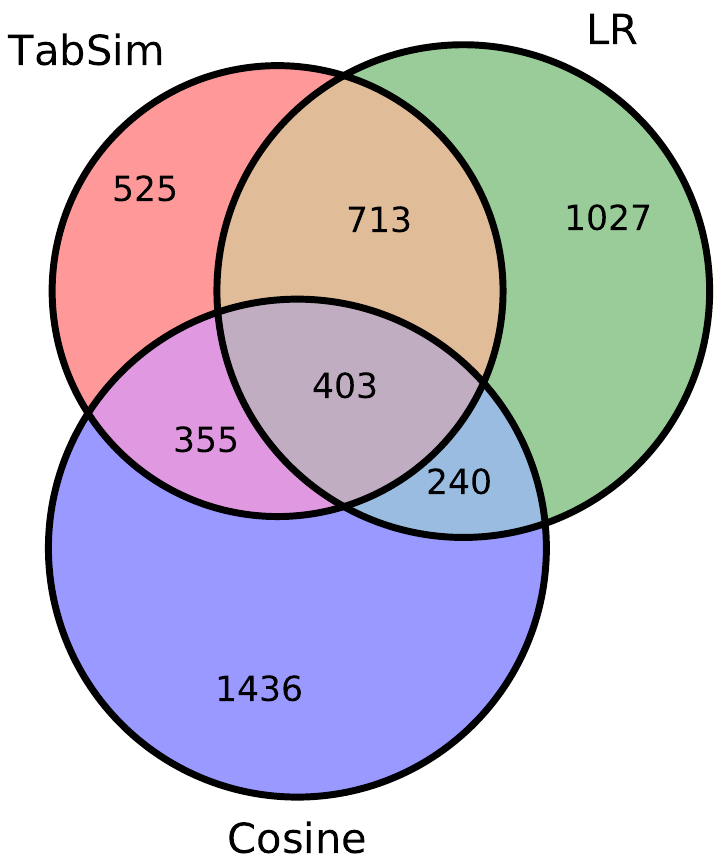}
\\
(d)&(e) &(f)\\
\end{tabular}
    \caption{Venn diagrams displaying the area of intersection among the FN sets \textit{TabSim}, \textit{LR} and \textit{Cosine} on PMC (a), arXiv (b) and Wikipedia (c) and their FP sets on PMC (d), arXiv  (e) and Wikipedia (f).}
\label{fig:venn}
\end{figure}

To better understand the results of the different methods and their differences, we computed the sets of false negative (FN) and false positive (FP) classifications (at default threshold of $0.5$) of \textit{TabSim}, \textit{LR} and \textit{Cosine} and show their mutual overlaps for each corpus as Venn diagram in Figure~\ref{fig:venn}. We restrict the analysis to \textit{LR} and \textit{Cosine} as best of their classes to keep the diagrams easy to read. The diagrams show that the errors produced by \textit{TabSim}, \textit{LR} and \textit{Cosine} are rather different in all cases but FN on PMC, where overlaps are larger. In the other fixed settings, \textit{TabSim} always has the least number of unique errors.

We manually examined the non-overlapping FNs (a-c) produced by each individual classifier on the PMC corpus. We observed that \textit{LR} and \textit{Cosine} have difficulty with the identification of similar table pairs containing genetic primers or gene names which are semantically similar but syntactically different (recall that PMC tables are randomly sampled from a corpus of biomedical publications). Although word embeddings are capable of learning semantic information in general, they cannot provide an appropriate representation for these tokens which are infrequently seen in documents. \textit{TabSim} suffers less from this problem as the embedding vectors are retrained during model training. However, compared to the other two methods \textit{TabSim} tends to misclassify table pairs with very long cells, i.e., many tokens per cell, which can be explained by \textit{TabSim}'s restriction to only consider a limited table cell size.

\subsection{Relation Ranking}
We compare \textit{TabSim} with three competitors and two baselines on three corpora regarding their capability of ranking tables given a query table in terms of NDCG$@k$ for $k = \{5,10\}$. 5F-CV NDCGs  are provided in Table~\ref{tab:ndcg}. \textit{TabSim} achieves the best values in three out of six settings and is rather close to the best in further two settings. It is clearly outperformed by \textit{RF} in the NDCG$@$5 score on the arXiv corpus while \textit{RF} classification performance is quite low (see Section~\ref{subsec:BCP}). Indeed, it places tables higher in the ranking while their similarity scores are lower. \textit{TabSim} is slightly worse than \textit{Cosine} and \textit{Google Fusion} on PMC for both $k$ values.
 Since in PMC for each query table the five tables with highest cosine
similarity were included, it is expected that methods also relying on
cosine similarity achieve very good results. We find it rather
reassuring that \textit{TabSim}, not using cosine similarity, achieves competitive results also under this setting.
 Consistent with Figure~\ref{fig:rocs}, all methods perform rather similar on PMC, while differences are much larger on the other two corpora. Averaged over the three corpora, \textit{TabSim} outperforms \textit{LR}, \textit{RF}, \textit{Cosine}, \textit{Google Fusion} and \textit{Jaccard} in terms of NDCG$@$10 by 3.0\% pp, 1.3\% pp, 17.2\% pp, 19.0\% pp and 15.8\% pp, respectively. \textit{TabSim} also outperforms all competitors in terms of NDCG$@$5 by at least 4.5\% pp, except \textit{RF}.

\begin{table}[ht!]
\def\arraystretch{.95}

\centering
\caption{5F-CV NDCGs (\%) for \textit{Jaccard}, \textit{Cosine}, \textit{Google Fusion}, \textit{RF}, \textit{LR} and \textit{TabSim} over three corpora. Best value per measure is in bold.}
\label{tab:ndcg}

\begin{tabular}{llcc}
\hline
\textbf{Corpora}&\textbf{Method}&\textbf{NDCG$\boldsymbol @$5}&\textbf{NDCG$\boldsymbol @$10} 
\\
\hline

\multirow{6}{*}{\textbf{PMC}}& \textbf{\textit{Jaccard}}& 93.10 & 94.66
\\

&\textbf{\textit{Cosine}}& \textbf{95.58} & \textbf{95.68}
\\
&\textbf{\textit{Google Fusion}}& 94.51 & 95.04
\\
&\textbf{\textit{RF}} & 90.53 & 92.03
\\
&\textbf{\textit{LR}} & 92.11 & 93.13
\\
&\textbf{\textit{TabSim}}&	 93.76 & 94.57
\\
\hline
\multirow{6}{*}{\textbf{arXiv}}& \textbf{\textit{Jaccard}}& 40.53 & 41.09
\\
&\textbf{\textit{Cosine}} & 35.03 & 36.18
\\
&\textbf{\textit{Google Fusion}}& 29.17 & 32.11
\\
&\textbf{\textit{RF}} & \textbf{81.07} & 82.26
\\
&\textbf{\textit{LR}} & 62.25 & 72.48
\\
&\textbf{\textit{TabSim}} & 74.15 & \textbf{82.71}
\\

\hline
\multirow{6}{*}{\textbf{Wikipedia}}& \textbf{\textit{Jaccard}} & 91.38 & 91.45
\\
&\textbf{\textit{Cosine}} & 91.06 & 91.14
\\
&\textbf{\textit{Google Fusion}} & 90.13 & 90.28
\\
&\textbf{\textit{RF}}&  96.46 & 96.50
\\
&\textbf{\textit{LR}}&  97.18 & 97.20
\\
&\textbf{\textit{TabSim}}&  \textbf{97.28} & \textbf{97.32}
\\
\hline
\end{tabular}
\end{table}

\section{Conclusion}
We presented \textit{TabSim}, a new method for assessing table similarity which uses Siamese neural networks to learn a similarity measure from a gold standard corpus of table pairs. We showed that, in comparison to five other methods of which three are also rooted in applications based on table similarity,  \textit{TabSim} attains considerably higher precision, recall, F1-score, and accuracy measures on three different corpora. Our results also demonstrate that, among different configurations of \textit{TabSim}, the model which uses self-attention neural networks achieve the highest performance, probably because it is, different from the 2d-based CNN or the sequence-based Bi-LSTM, invariant to row or column permutations. As part of our research, we also created the first specific gold standard corpus for table similarity research, containing 1500 table pairs manually scored regarding their semantic similarity. Although the corpus was created in a way that gives methods relying on cosine similarity a competitive advantage, \textit{TabSim} also leads the field on this corpus. 

A disadvantage of \textit{TabSim} is its high execution time; it takes, on average, about 5 ms to classify a pair of tables (compared to 2 ms for \textit{LR} and 0.5 ms for \textit{Cosine}). This high runtime is certainly not appropriate when using \textit{TabSim} as a similarity function for a table similarity search engine, where the query table would be compared to every table from the corpus at search time. In future work, we plan to focus on designing scalable table search engines that use \textit{TabSim} at the core but apply additional techniques for early search space pruning.



\bibliographystyle{IEEEtran}

\bibliography{acl2019}

\end{document}